\documentclass[english]{sobraep}
\usepackage{kotex}
\usepackage{amsmath}
\usepackage{color}
\usepackage{romannum}
\usepackage{placeins}
\usepackage{float}
\usepackage{verbatim}
\usepackage{amssymb}
\usepackage{multirow}
\usepackage{algorithm}
\usepackage{algorithmicx, algpseudocode}
\usepackage{multirow}

\title{Multitask Emotion Recognition Model with Knowledge Distillation and Task Discriminator}

\author{Euiseok Jeong$^{1}$, Geesung Oh$^{1}$, Sejoon Lim$^{2}$\\
	\normalsize $^{1}$Graduate School of Automotive Engineering, Kookmin University, Seoul\\ \{euiseok\_jeong, gsethan17\}@ kookmin.ac.kr\\
	\normalsize $^{2}$Department of Automobile and IT Convergence, Kookmin University, Seoul\\ lim@kookmin.ac.kr
}

\begin{document}
\maketitle
\begin{abstract}
Due to the collection of big data and the development of deep learning, research to predict human emotions in the wild is being actively conducted.
We designed a multi-task model using ABAW dataset to predict valence-arousal, expression, and action unit through audio data and face images at in real world.
We trained model from the incomplete label by applying the knowledge distillation technique.
The teacher model was trained as a supervised learning method, and the student model was trained by using the output of the teacher model as a soft label.
As a result we achieved 2.40 in Multi Task Learning task validation dataset.
\end{abstract}

\section{Introduction}
\label{sec:Intro}

Human affective behavior research is an important field in the human computer interaction field, and is actively studied with the development of big data and deep learning technologies.
However, for several reasons, it is difficult to apply the effective behavioral study.
To address there problems, 3rd Affective Behavior Analysis in-the-wild (ABAW) Competition is held in conjunction with IEEE International Conference on Computer Vision and Pattern Recognition (CVPR), 2022.
3rd ABAW includes Valence-Arousal (VA) Estimation Challenge, Expression (Expr) Classification Challenge, Action Unit (AU) Detection Challenge, Multi-Task-Learning (MTL) Challenge.
146 teams participated in the 2nd ABAW to improve the performance of the human emotion model in the real world.\\
We participated in the 2nd ABAW valence-arousal (VA) Estimation Challenge and ranked 10th with a average CCC(concordance correlation) of valence and arousal 0.261 with CAPNet\cite{oh2021causal} that predicted the future using only past image data.
We proposed a multitasking model that improves performance. Improvement points are as follows.\\
First, audio input stream was added. 
Audio information is closely related to human emotions. It may reveal human emotions in the form of speech \cite{ooi2014new, khalil2019speech}, or elicitate human emotions in the form of music \cite{bhatti2016human} and sound \cite{weninger2013acoustics}. Kuhnke et al. \cite{kuhnke2020two} and Deng et al. \cite{deng2021iterative} shows that audio input is effective for recognizing human emotions through the ablation study.
Accordingly, audio input for past 10 seconds was added. The audio feature is extracted from a audio input via SoundNet\cite{aytar2016soundnet}.\\
Second, knowledge distillation proposed by Hinton et al\cite{hinton2015distilling} was applied to improve generalization performance. It is a technique that improves generalization performance of a student model by transferring dark knowledge to the student model by utilizing the output of the pre-trained teacher model as a soft label. We trained the teacher model using only the data with ground truths on the task. After training the teacher model, output of the teacher model is used as a soft label to transfer dark knowledge to the student model.

Third, we extracted task-wise features by adding gradient reverse layer and task discriminator.
The feature extracted from the feature extractor was made to be task-independent by gradient reversal layer and task discriminator.

As a result of applying the above methods, we achieved 2.40 performances for the 2022 ABAW validation set, which exceeded baseline performance 0.3 proposed by \cite{kollias2022abaw}.

Primary contributions of this paper are:
\begin{itemize}
\item 
We designed a multi-task model to simultaneously perform Valence-Arousal estimation, Expression Classification, and Action Unit Detection through Audio and image multi input.
\item 
By applying the knowledge distillation technique, we improved the generalization performance of the multi-task model.

\item 
We extracted task-independent features by adding task discriminator and Gradient Reverse Layer.

\end{itemize}

\section{Related work}

\subsection{ABAW}
Recent years, data-based research on human affective behavior research is growing rapidly, and the ABAW competition is contributing greatly to that foundation. ABAW which has been held by Kollias et al. for 3 years, is a competition that predicts the affect of characters and competes for their prediction performance using video of people appearing. \cite{kollias2022abaw, kollias2021analysing, kollias2020analysing, kollias2019face, kollias2021distribution, kollias2019deep, kollias2019expression, kollias2021affect, zafeiriou2017aff}. The first competition was held at 2020 International Conference on Automatic Face and Gesture Recognition (FG), the second at 2021 International Conference on Computer Vision (ICCV), and the third at CVPR. Basically ABAW competition consists of three challenges: 2 dimensional affect classification, 7 categorical affect classification, and 12 facial action unit detection. In the first and second competitions, the top-ranked teams all proposed deep learning-based recognition models, and most of top-ranked models that recognize three challenges simultaneously \cite{kuhnke2020two, deng2020multitask, youoku2020multi, deng2021iterative, zhang2021prior, zhang2021continuous, youoku2021multi}. From the third competition, MTL challenge for multi-label models that recognize three challenges at the same time is newly established, and the existing three challenges are limited to uni-label models. In addition to the output, the top-ranked teams are classified by the input data. All top-ranked teams use images from video data, but there are teams that use only image data \cite{deng2020multitask, youoku2020multi, zhang2021prior, youoku2021multi}, and teams that use audio data along with image data to improve performance \cite{kuhnke2020two, deng2021iterative, zhang2021continuous}.

\subsection{Knowledge distillation}
The knowledge distillation was proposed by hinton et al\cite{hinton2015distilling}.
The output of the pretrained teacher model is scaled by softmax function with temperature and used as a soft label to train the student model.
The student model learns inter-class similarity(dark knowledge) from the soft label of the teacher model and achieves performance similar to the teacher model despite being shallower than the teacher model.
Zhang et al.\cite{zhang2019your} also improved performance through a self-distillation technique that configured the structure of the teacher model and the student model equally.
Many teams used the knowledge distillation model at the 2021 ABAW and were also on the leaderboard.
In particular, Deng et al.\cite{deng2021iterative} made it possible to train deeper dark knowledge using the knowledge distillation technique, the ensemble technique, and the generation technique in which a trained student model becomes a teacher model and trains a new student model.

\subsection{Domain adaptation}
Domain adaptation techniques proposed by Ganin et al.\cite{ganin2015unsupervised} to improve accuracy in target domains without ground truth through data from source domains with ground truth.
Gradient Reverse Layer(GRL), which is multiply -1 to gradient when performing back propagation, and a domain discriminator are added to allow the feature extractor to extract features independent of the domain.

\section{Problem definition}
\label{sec:problem}
\{$X$,$Y$\} denoted train data. $X$ represents input data and $Y$ represents ground truths.
Model function inputs $X$ and outputs $Y$.
For the convenience of notation, it is assumed that the batch size of all tasks is $N$, but even if the number of tasks' data is different, it can be configured to be divided by the same number of iterations.
$X$ consists of $X_{img}$ and $X_{aud}$. $X_{img}$ represents the facial image and $X_{aud}$ represents the audio data: 
$X=\{X_{img}\in R^{N\times n_{img} \times H\times H\times C_{img}},  {X_{aud}\in R^{N\times sr\cdot t_{aud}\times C_{aud}}}\}$.

$n_{img}$ is the number of input images. $H$ is width and height of the image and $C_{img}$ refers to the number of channel of image.
In audio data, $sr$ refers to sample rate of audio data. $t_{aud}$ means audio data time before prediction time. $C_{aud}$ is the number of channel of audio.
In this dataset, $n_{hz}$, frame rate of a video in dataset is 30, $sr$ is 22050 amd $C_{aud}$ is 2.
$X_{img}$ contains a image for the past $t_{img}$ seconds from the time of prediction. Of the total ${t_{img}\cdot n_{hz}}$ images, we extracted $n_i$ images with stride $s$.
$H$ is 112 and $C_{img}$ is 3 in this dataset.

The ground truths $Y$ consists of four types: $Y=\{Y_{VA}\in R^{N\times2},  Y_{EXPR}\in R^{N\times8}, Y_{AU}\in R^{N\times12}, Y_{MTL}\in R^{N\times22}\}$.
$Y_{VA}$ represents a continuous valence and arousal in the range of [-1, 1].
$Y_{EXPR}$ is a one-hot encoded vector for eight categories of emotions: Neutral, Anger, Disgust, Fear, Happiness, Sadness, Surprise, and Other.
$Y_{AU}$ includes 12 facial action units labels: AU1, AU2, AU4, AU6, AU7, AU10, AU12, AU15, AU23, AU24, AU25 and AU26.
Model function is denoted by $f$.
In teacher-student train algorithm, teacher model denoted by $f^{tea}$ and soft label from teacher model in task $i$ is $f_i^{tea}(X)$.

\section{Methodologies}

\subsection{Architecture}
\begin{figure}[h]
% \widefigure
  \centering
  \includegraphics[scale=0.30]{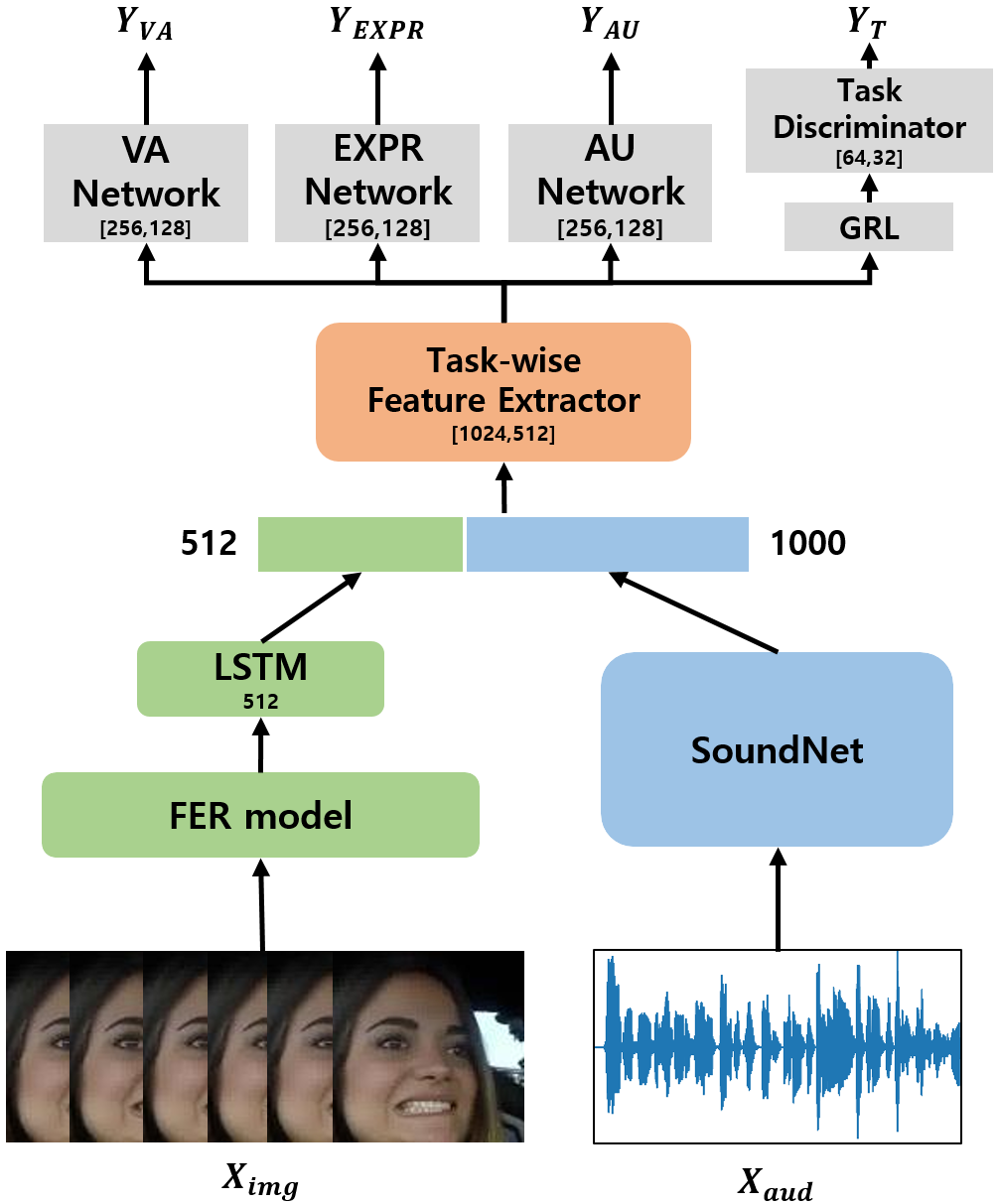}
  \caption{Model architecture}
  \label{fig:model_architecture}
\end{figure}
\label{sec:Architecture}
We used CNN architecture based on FER model of DRER\cite{oh2021drer} to extract image representation.
It aims to predict the valence and arousal of the driver.
The FER model is composed of ResNeXt\cite{} and SENet\cite{} and is pre-trained with AffectNet\cite{mollahosseini2017affectnet}.
A detailed description of the FER model can be found in \cite{oh2021drer}.
Output of FER model from single image is a 512-dimensional vector.
As mentioned in section \ref{sec:problem}, Six images are fed into the FER model and (6,512) shaped image representation is extracted.
The extracted image representation is fed into the LSTM model to capture the temporal feature.\\
% SoundNet
To extract sound representation, we adopted SoundNet proposed by Aytar et al.\cite{aytar2016soundnet}. 
SoundNet composed of a series of one-dimensional convolutions. The audio data for $t_{aud}$ seconds is fed into SoundNet and extracted sound representation from SoundNet is the 1000-dimensional vector.

SoundNet is trained to transfer visual knowledge through well established visual recognition models, leveraging massive amounts of unlabeled videos \cite{aytar2016soundnet}. Therefore, it is particularly effective in the cross-model approach by using both video and audio \cite{kajihara2017imaginary}.

 Sound representation and image representation are concatenated before fed into feature extractor.
The feature extractor consists of a dense module containing linear layer, swish activation function, batch normalization, and dropout layer. 
Task-wise feature is extracted from feature extractor.
% Each task has their own task model. 
Each task has their own task model.
The task-wise feature is fed into each task model, and each task model outputs prediction for each task.
The task-wise feature is fed into the task discriminator via the gradient reversal layer. The task discriminator outputs an output for which task is.
When backpropagation is performed using the loss between the task prediction value and the task label value, the gradient reversal layer multiplies the gradient by -1.
The feature extractor is trained in a direction in which the task discriminator cannot distinguish task from feature, and at the same time, it is trained to improve the performance of each task network.

\subsection{Loss function}
\label{sec:loss}
When batch data has a ground truth for the task, we calculate the loss between the output of the model and the ground truth. We refers to this as \emph{supervision loss}.
When batch data does not have a ground truth for the task, we calculate the loss between the output of the teacher model called soft label and the output of the student model. We refers to this as \emph{distillation loss}.

\paragraph{Supervision loss}
For valence-arousal estimation task, we used Concordance Correlation Coefficient (CCC) loss.
CCC is defined as follows:
\begin{equation}\label{ccc}
CCC(y,\hat{y})=\frac{2\rho \sigma_{y}\sigma_{\hat{y}}}{\sigma_{y}^{2}+\sigma_{\hat{y}}^{2}+{(\mu_{y}-\mu_{\hat{y}}})^2}
\end{equation}
where, $y$ is ground truth and $\hat{y}$ is prediction value. and $\sigma$ and $\mu$ is standard deviation and means computed over the batch. 
The supervision loss for valence-arousal estimation task is as follows:
\begin{equation}\label{ccc_loss}
L_{VA}^{S}=1-CCC(Y_{VA},f_{VA}^{tea}(X))
\end{equation}

For expression classification task, we use cross entropy loss.
cross entropy defined as follows:
\begin{equation}\label{ce}
CE(y,\hat{y})=-\sum_{c=1}^{C}ylog(\hat{y})
\end{equation}
$C$ is the number of classes.
The supervision loss for expression classification task is as follows:
\begin{equation}\label{expr_loss}
L_{EXPR}^{S}=CE(Y_{EXPR},f_{EXPR}^{tea}(X))
\end{equation}
For AU detection task, we use binary cross entropy. Binary cross entropy defined as follows:
\begin{equation}\label{bce}
BCE(y,\hat{y})=[ylog(\hat{y})+(1-y)log(1-\hat{y})]
\end{equation}
The supervision loss for AU detection task $L_{AU}^S$ is defined as follows:
\begin{equation}\label{au_loss}
L_{AU}^{S}=BCE(Y_{AU},f_{AU}^{tea}(X))
\end{equation}
The supervision loss for MTL task $L_{MTL}^S$ is sum of each task loss:
\begin{equation}\label{mtl_loss}
L_{MTL}^S=L_{VA}^S+L_{EXPR}^S+L_{AU}^S
\end{equation}

\paragraph{Distillation loss}
To transfer the dark knowledge of the teacher model to the student model, the loss between the output of the teacher model and the output of the student model is calculated.
The distillation loss equation for each task is as follows.
\begin{equation}\label{va_distil}
L_{VA}^{D}=1-CCC(f_{VA}^{stu}(X),f_{VA}^{tea}(X))
\end{equation}
\begin{equation}\label{expr_distil}
L_{EXPR}^{D}=CE(f_{EXPR}^{stu}(X),f_{EXPR}^{tea}(X))
\end{equation}
\begin{equation}\label{au_distil}
L_{AU}^{D}=BCE(f_{AU}^{stu}(X),f_{AU}^{tea}(X))
\end{equation}
MTL tasks have labels for all tasks, so distillation loss of MTL task are not calculated.

\paragraph{Task classification loss}
% Task discriminator에서 나온 output과 VA,EXPR,AU task에 대해 one-hot encoded된 task label사이에 cross entropy loss를 구함.
Task classification loss is the cross entropy loss between the output from the Task discriminator and the task label one-hot encoded for VA, EXPR, and AU tasks. Task classification loss is defined as follows:
\begin{equation}\label{task_classification_loss}
L^T = CE(f^{task}(X), Y_{task})
\end{equation}

\paragraph{Train loss}
The teacher loss of task $i$ is defined using supervision loss and task classification loss.
\begin{equation}\label{teacher_loss}
Loss_i^{tea}=\sum_{n=1}^{N}(L_i^S + L_i^T)
\end{equation}

Student loss of task $i$ is defined as follows by combining supervision loss, distillation loss and task classification loss.
\begin{equation}\label{student_loss}
Loss_i^{stu}=\sum_{n=1}^{N}\{\gamma_i\cdot(\alpha\cdot L_i^S +L_i^D) +\delta \cdot  L_i^T +\sum_{j\neq i}\beta \cdot \gamma_j \cdot L_i^D\}
\end{equation}

Where, 
Gamma is the task weight for each task proposed by deng et al\cite{deng2021iterative}.
While training, the number of epochs that validation performance is not improved for each task is counted.
The more counts, the greater the weight for task loss to boost training.
When the counted number of task $i$ is $n_i$, the weight of that task loss is $\gamma_i = e^{0.5n_i}$.
$\alpha$ is the hyperparameter of the weight between supervision loss and distillation loss for tasks with a ground truth label.
$\beta$ is a hyperparameter representing the weight of distillation losses for tasks without a ground truth label.
$\delta$ refers to weight of task classification loss

\subsection{Train procedure}
\label{sec:train_procedure}
The teacher model trains to minimize equation (\ref{teacher_loss}) consisting of supervision loss and task classification loss without distillation loss.
After the training of the teacher model is completed, the student model trains to minimize the equation (\ref{student_loss}) using soft label from teacher model.
Train procedure of the teacher model and student model is described in algorithm 2.\ref{alg:train_alge}
\begin{algorithm}
\caption{Train procedure}
\label{alg:train_alge}
\begin{algorithmic}[0]
    \Require 
    \State parameters: $\theta_t$(teacher), $\theta_s$(student)
    \State Epoch: $N$
    \State Task: $[VA,EXPR,AU,MTL]$
    \State learning rate: $lr$\\
    
    \State \textbf{Train} teacher model
        \State $n_{epoch}$ = 0
            \While{$n_{epoch}$ < $N$} 
                \While{not epoch end}
                    \For{$i \in Task$}
                        \If{$i\ is\ MTL$}
                            \State $loss = L_{MTL}^S$
                        \Else
                            \State $loss = Loss_i^{tea}$
                        \EndIf
                        \State $\theta_t\leftarrow\theta_t-lr\cdot\frac{\partial loss}{\partial \theta_t}$
                    \EndFor
                \EndWhile
                \State $n_{epoch}\leftarrow n_{epoch}+1$
            \EndWhile
            \\
    \State \textbf{Train} student model
        \State $n_{epoch}$ = 0
            \While{$n_{epoch}$ < $N$} 
                \While{not epoch end}
                    \For{$i \in Task$}
                        \If{$i\ is\ MTL$}
                            \State $loss = L_{MTL}^S$
                        \Else
                            \State $loss = Loss_i^{stu}$
                        \EndIf
                        \State $\theta_s\leftarrow\theta_s-lr\cdot\frac{\partial loss}{\partial \theta_s}$
                    \EndFor
                \EndWhile
                \State $n_{epoch}\leftarrow n_{epoch}+1$
            \EndWhile

\end{algorithmic}
\end{algorithm}
\section{Experiments}

\subsection{Experiments setting}
We used Adam optimizer, the learning rate was set to 0.0001, and the batch size was 256.
We train for 20 epochs and stop training if there is no improvement in validation performance for 5 epochs.
We set $t_{aud}$ to 10, $t_{img}$ to 2, and $s$ to 6.
In Loss function, $\alpha$ was set to 10, and $\beta$ was set to 0.9. $\delta$ was set to 1 or started with 0 and increased linearly to 1 as train progressed.
$t$ was set to 2.5.
In feature extractor layer, 0.5 random dropout was applied.
$\delta$ was set to 0.5 or 1, and started with 0 and increased linearly to 1 as train progressed.
\subsection{Metrics}
We used metrics suggested in \cite{kollias2022abaw} to evaluate model performance.
The metric of Multi-Task-Learning is the sum of the metrics of valence-arousal, expression, and action unit challenge.
The valence-arousal metric is a concordance correlation coefficient (CCC), and the metric of expression recognition is F1 score across all 8 categories (i.e. macro F1 score). 
The metric of the action unit detection is the average F1 score across all 12 AUs (i.e. macro F1 score).
Although we are participating in the multi task learning challenge, validation was also performed on valence-arousal, expression classification, and action unit detection tasks.
\subsection{Results}

We evaluated the teacher and student model to which the knowledge distillation technique was applied using the ABAW validation set and compared it with the model to which the task discriminator was added.
% $\delta$이 1인 모델과 0부터 train이 진행되며 1로 linear하게 증가하는 모델의 성능을 비교하였다.
% validation 결과는 \textcolor{red}{실험 결과표 ref}에 described 되어있다.
Validation results are described in TABLE \ref{validation_result}.

In all models, it can be seen that the Student model is outperformed than the Teacher model.
We can see that the dark knowledge transferred by the Teacher model helped improve generalization performance.
Domain discriminator
However, validation performance tended to decrease when the Domain discriminator was added.
It is possible that feature extractor has already extracted task wise feature in the learning process, and task classification loss has had a negative effect on task performance.

\begin{table*}[]
\centering
\caption{Validation performance of teacher, student model.}
\label{validation_result}
\begin{tabular}{|c|cccc|}
\hline
Model    & \multicolumn{4}{c|}{Performance(VA/EXPR/AU/MTL)}                                                                                                     \\ \hline
$\delta$ & \multicolumn{1}{c|}{-}                   & \multicolumn{1}{c|}{0.5}                 & \multicolumn{1}{c|}{1}                   & 0$\rightarrow$1     \\ \hline
Teacher  & \multicolumn{1}{c|}{0.64/0.56/0.51/2.08} & \multicolumn{1}{c|}{0.58/0.51/0.47/1.76} & \multicolumn{1}{c|}{0.59/0.46/0.47/1.81} & 0.61/0.41/0.39/1.69 \\ \hline
Student  & \multicolumn{1}{c|}{\textbf{0.63/0.57/0.51/2.40}} & \multicolumn{1}{c|}{0.63/0.58/0.52/2.38} & \multicolumn{1}{c|}{0.61/0.53/0.50/2.36} & 0.59/0.51/0.51/2.24 \\ \hline
\end{tabular}
\end{table*}

\section{Conclusions}
In this paper, we proposed a multi stream, multi task model applying knowledge distillation to improve the generalization performance by training with the incomplete label.
We tried to extract task independent features by adding a Gradient reversal layer and a task discriminator.
In the future, We will study minimizing input data time without performance decline, and further study how to improve performance through the application of task discriminator.

\bibliographystyle{unsrt}
\bibliography{main}

\begin{thebibliography}{10}

\bibitem{oh2021causal}
Geesung Oh, Euiseok Jeong, and Sejoon Lim.
\newblock Causal affect prediction model using a past facial image sequence.
\newblock In {\em Proceedings of the IEEE/CVF International Conference on
  Computer Vision}, pages 3550--3556, 2021.

\bibitem{ooi2014new}
Chien~Shing Ooi, Kah~Phooi Seng, Li-Minn Ang, and Li~Wern Chew.
\newblock A new approach of audio emotion recognition.
\newblock {\em Expert systems with applications}, 41(13):5858--5869, 2014.

\bibitem{khalil2019speech}
Ruhul~Amin Khalil, Edward Jones, Mohammad~Inayatullah Babar, Tariqullah Jan,
  Mohammad~Haseeb Zafar, and Thamer Alhussain.
\newblock Speech emotion recognition using deep learning techniques: A review.
\newblock {\em IEEE Access}, 7:117327--117345, 2019.

\bibitem{bhatti2016human}
Adnan~Mehmood Bhatti, Muhammad Majid, Syed~Muhammad Anwar, and Bilal Khan.
\newblock Human emotion recognition and analysis in response to audio music
  using brain signals.
\newblock {\em Computers in Human Behavior}, 65:267--275, 2016.

\bibitem{weninger2013acoustics}
Felix Weninger, Florian Eyben, Bj{\"o}rn~W Schuller, Marcello Mortillaro, and
  Klaus~R Scherer.
\newblock On the acoustics of emotion in audio: what speech, music, and sound
  have in common.
\newblock {\em Frontiers in psychology}, 4:292, 2013.

\bibitem{kuhnke2020two}
Felix Kuhnke, Lars Rumberg, and J{\"o}rn Ostermann.
\newblock Two-stream aural-visual affect analysis in the wild.
\newblock In {\em 2020 15th IEEE International Conference on Automatic Face and
  Gesture Recognition (FG 2020)}, pages 600--605. IEEE, 2020.

\bibitem{deng2021iterative}
Didan Deng, Liang Wu, and Bertram~E Shi.
\newblock Iterative distillation for better uncertainty estimates in multitask
  emotion recognition.
\newblock In {\em Proceedings of the IEEE/CVF International Conference on
  Computer Vision}, pages 3557--3566, 2021.

\bibitem{aytar2016soundnet}
Yusuf Aytar, Carl Vondrick, and Antonio Torralba.
\newblock Soundnet: Learning sound representations from unlabeled video.
\newblock {\em Advances in neural information processing systems}, 29, 2016.

\bibitem{hinton2015distilling}
Geoffrey Hinton, Oriol Vinyals, Jeff Dean, et~al.
\newblock Distilling the knowledge in a neural network.
\newblock {\em arXiv preprint arXiv:1503.02531}, 2(7), 2015.

\bibitem{kollias2022abaw}
Dimitrios Kollias.
\newblock Abaw: Valence-arousal estimation, expression recognition, action unit
  detection \& multi-task learning challenges.
\newblock {\em arXiv preprint arXiv:2202.10659}, 2022.

\bibitem{kollias2021analysing}
Dimitrios Kollias, Irene Kotsia, Elnar Hajiyev, and Stefanos Zafeiriou.
\newblock Analysing affective behavior in the second abaw2 competition.
\newblock {\em arXiv preprint arXiv:2106.15318}, 2021.

\bibitem{kollias2020analysing}
Dimitrios Kollias, Attila Schulc, Elnar Hajiyev, and Stefanos Zafeiriou.
\newblock Analysing affective behavior in the first abaw 2020 competition.
\newblock In {\em 2020 15th IEEE International Conference on Automatic Face and
  Gesture Recognition (FG 2020)}, pages 637--643. IEEE, 2020.

\bibitem{kollias2019face}
Dimitrios Kollias, Viktoriia Sharmanska, and Stefanos Zafeiriou.
\newblock Face behavior a la carte: Expressions, affect and action units in a
  single network.
\newblock {\em arXiv preprint arXiv:1910.11111}, 2019.

\bibitem{kollias2021distribution}
Dimitrios Kollias, Viktoriia Sharmanska, and Stefanos Zafeiriou.
\newblock Distribution matching for heterogeneous multi-task learning: a
  large-scale face study.
\newblock {\em arXiv preprint arXiv:2105.03790}, 2021.

\bibitem{kollias2019deep}
Dimitrios Kollias, Panagiotis Tzirakis, Mihalis~A Nicolaou, Athanasios
  Papaioannou, Guoying Zhao, Bj{\"o}rn Schuller, Irene Kotsia, and Stefanos
  Zafeiriou.
\newblock Deep affect prediction in-the-wild: Aff-wild database and challenge,
  deep architectures, and beyond.
\newblock {\em International Journal of Computer Vision}, 127(6):907--929,
  2019.

\bibitem{kollias2019expression}
Dimitrios Kollias and Stefanos Zafeiriou.
\newblock Expression, affect, action unit recognition: Aff-wild2, multi-task
  learning and arcface.
\newblock {\em arXiv preprint arXiv:1910.04855}, 2019.

\bibitem{kollias2021affect}
Dimitrios Kollias and Stefanos Zafeiriou.
\newblock Affect analysis in-the-wild: Valence-arousal, expressions, action
  units and a unified framework.
\newblock {\em arXiv preprint arXiv:2103.15792}, 2021.

\bibitem{zafeiriou2017aff}
Stefanos Zafeiriou, Dimitrios Kollias, Mihalis~A Nicolaou, Athanasios
  Papaioannou, Guoying Zhao, and Irene Kotsia.
\newblock Aff-wild: valence and arousal'in-the-wild'challenge.
\newblock In {\em Proceedings of the IEEE conference on computer vision and
  pattern recognition workshops}, pages 34--41, 2017.

\bibitem{deng2020multitask}
Didan Deng, Zhaokang Chen, and Bertram~E Shi.
\newblock Multitask emotion recognition with incomplete labels.
\newblock In {\em 2020 15th IEEE International Conference on Automatic Face and
  Gesture Recognition (FG 2020)}, pages 592--599. IEEE, 2020.

\bibitem{youoku2020multi}
Sachihiro Youoku, Yuushi Toyoda, Takahisa Yamamoto, Junya Saito, Ryosuke
  Kawamura, Xiaoyu Mi, and Kentaro Murase.
\newblock A multi-term and multi-task analyzing framework for affective
  analysis in-the-wild.
\newblock {\em arXiv preprint arXiv:2009.13885}, 2020.

\bibitem{zhang2021prior}
Wei Zhang, Zunhu Guo, Keyu Chen, Lincheng Li, Zhimeng Zhang, and Yu~Ding.
\newblock Prior aided streaming network for multi-task affective recognitionat
  the 2nd abaw2 competition.
\newblock {\em arXiv preprint arXiv:2107.03708}, 2021.

\bibitem{zhang2021continuous}
Su~Zhang, Yi~Ding, Ziquan Wei, and Cuntai Guan.
\newblock Continuous emotion recognition with audio-visual leader-follower
  attentive fusion.
\newblock In {\em Proceedings of the IEEE/CVF International Conference on
  Computer Vision}, pages 3567--3574, 2021.

\bibitem{youoku2021multi}
Sachihiro Youoku, Takahisa Yamamoto, Junya Saito, Akiyoshi Uchida, Xiaoyu Mi,
  Ziqiang Shi, Liu Liu, Zhongling Liu, Osafumi Nakayama, and Kentaro Murase.
\newblock Multi-modal affect analysis using standardized data within subjects
  in the wild.
\newblock {\em arXiv preprint arXiv:2107.03009}, 2021.

\bibitem{zhang2019your}
Linfeng Zhang, Jiebo Song, Anni Gao, Jingwei Chen, Chenglong Bao, and Kaisheng
  Ma.
\newblock Be your own teacher: Improve the performance of convolutional neural
  networks via self distillation.
\newblock In {\em Proceedings of the IEEE/CVF International Conference on
  Computer Vision}, pages 3713--3722, 2019.

\bibitem{ganin2015unsupervised}
Yaroslav Ganin and Victor Lempitsky.
\newblock Unsupervised domain adaptation by backpropagation.
\newblock In {\em International conference on machine learning}, pages
  1180--1189. PMLR, 2015.

\bibitem{oh2021drer}
Geesung Oh, Junghwan Ryu, Euiseok Jeong, Ji~Hyun Yang, Sungwook Hwang, Sangho
  Lee, and Sejoon Lim.
\newblock Drer: Deep learning--based driver’s real emotion recognizer.
\newblock {\em Sensors}, 21(6):2166, 2021.

\bibitem{mollahosseini2017affectnet}
Ali Mollahosseini, Behzad Hasani, and Mohammad~H Mahoor.
\newblock Affectnet: A database for facial expression, valence, and arousal
  computing in the wild.
\newblock {\em IEEE Transactions on Affective Computing}, 10(1):18--31, 2017.

\bibitem{kajihara2017imaginary}
Yuma Kajihara, Shoya Dozono, and Nao Tokui.
\newblock Imaginary soundscape: Cross-modal approach to generate pseudo sound
  environments.
\newblock In {\em Proceedings of the Workshop on Machine Learning for
  Creativity and Design (NIPS 2017), Long Beach, CA, USA}, pages 1--3, 2017.

\end{thebibliography}

\end{document}